\documentclass[10pt,twocolumn,letterpaper]{article}

\usepackage{iccv}
\usepackage{times}
\usepackage{epsfig}
\usepackage{graphicx}
\usepackage{amsmath}
\usepackage{amssymb}
\usepackage{caption}
\usepackage{enumitem}
\usepackage{booktabs}

\usepackage{multicol}
\usepackage{caption}
\usepackage{makecell}
\usepackage{multirow}
\usepackage{color, colortbl}
\usepackage{cuted}
\usepackage{capt-of}
\usepackage[numbers,sort,compress]{natbib}

\definecolor{Gray}{gray}{0.9}
\newcommand{\modelname}[0]{DSR\xspace}

\usepackage[pagebackref=true,breaklinks=true,letterpaper=true,colorlinks,bookmarks=false]{hyperref}

\iccvfinalcopy %

\def\iccvPaperID{4330} %

\begin{document}

\title{Learning to Regress Bodies from Images using \\ Differentiable Semantic Rendering}

\author{
    Sai Kumar Dwivedi$^{1}$\quad \; 
    Nikos Athanasiou$^{1}$\quad \; 
    Muhammed Kocabas$^{1,2}$\quad \; 
    Michael J. Black$^{1}$\\
    $^1$Max Planck Institute for Intelligent Systems, T\"{u}bingen, Germany \quad 
    $^2$ETH Zurich\\
    \normalsize \texttt{\{\href{mailto:sdwivedi@tue.mpg.de}{sdwivedi}, \href{mailto:nathanasiou@tue.mpg.de}{nathanasiou}, \href{mailto:mkocabas@tue.mpg.de}{mkocabas}, \href{mailto:black@tue.mpg.de}{black}\}@tue.mpg.de}}
    
\maketitle
\ificcvfinal\thispagestyle{empty}\fi
\begin{strip}
 		\centering
 		\includegraphics[width=\textwidth,clip]{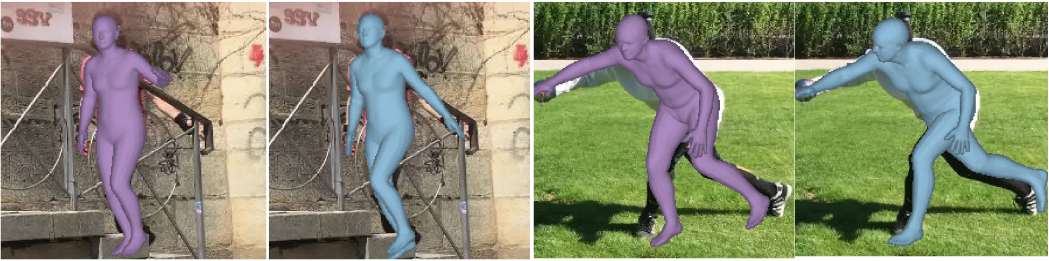}
  		\captionof{figure}{\textbf{Differentiable Semantic Rendering (DSR)}. A state-of-the-art approach~\cite{eft} (purple) fails to estimate accurate 3D pose and shape for in-the-wild scenarios. We address this by exploiting the clothing semantics of the human body. Our approach, \modelname,  (blue) captures more accurate 3D pose and shape compared to previous work.}
 		\label{fig:teaser}
\end{strip}

\begin{abstract}
Learning to regress 3D human body shape and pose (e.g.~SMPL parameters) from monocular images typically exploits losses on 2D keypoints, silhouettes, and/or part-segmentation when 3D training data is not available. 
Such losses, however, are limited because
2D keypoints do not supervise body shape and segmentations of people in clothing do not match projected minimally-clothed SMPL shapes.
To exploit richer image information about clothed people, we introduce higher-level semantic information about clothing to penalize clothed and non-clothed regions of the human body differently.
To do so, we train a body regressor using a novel ``\textbf{D}ifferentiable \textbf{S}emantic \textbf{R}endering \textbf{(\modelname)}'' loss. 
For Minimally-Clothed (MC) regions, we define the \modelname-MC loss, which encourages a tight match between a rendered SMPL body and the minimally-clothed regions of the image.
For clothed regions, we define the \modelname-C loss to encourage the rendered SMPL body to be inside the clothing mask. To ensure end-to-end differentiable training, we learn a semantic clothing prior for SMPL vertices from thousands of clothed human scans. We perform extensive qualitative and quantitative experiments to evaluate the role of clothing semantics on the accuracy of 3D human pose and shape estimation. We outperform all previous state-of-the-art methods on 3DPW and Human3.6M and obtain on par results on MPI-INF-3DHP. Code and trained models are available for research at \url{https://dsr.is.tue.mpg.de/}.
\end{abstract}

\section{Introduction}
Estimating 3D human pose and shape from in-the-wild images has received great research interest~\cite{smplify,eft, hmr,spin,Lassner2017UniteTP,nbf,pavlakos2018,densecor} because of its varied applications in animation, games, and the fashion industry.
One aspect that makes this problem challenging is the difficulty of obtaining accurate 3D ground-truth annotations, as they require either specialized --mostly indoors-- MoCap systems or careful calibration and setup of IMU sensors~\cite{vonMarcard2018}. 
Such data would facilitate training robust regressors paving the way for estimating human-scene interaction with greater granularity.

Given the lack of in-the-wild 3D ground-truth, the vast majority of previous methods focus on 2D keypoints~\cite{smplify, eft} with some learned 3D priors. %
Even though sparse 2D keypoints give useful constrained, relying only on these leads to unrealistic poses because of depth ambiguities and occlusion. 
They also do not provide reliable information about body shape.
On the other hand, relying too strongly on 3D priors introduces bias.
To circumvent this problem, recent approaches~\cite{nbf,pavlakos2018, denserac} propose to use part-segmentations or silhouettes. However, there is a mismatch between part-segmentations/silhouettes and projected SMPL bodies since segmentation covers clothed bodies while the common 3D body models are minimally clothed.
We propose an alternative approach to compensate for limited 3D supervision that leverages high-level 2D image cues.

Specifically, we propose more detailed clothing segmentation labels to supervise a neural network. Traditional multi-class clothing segmentation approaches cannot be directly applied as the segmentation loss tries to exactly match the rendered SMPL body. Hence, to make use of such labels, we need to reason about which parts of the SMPL body model correspond to which clothing label. This is non-trivial to obtain because a body part can be covered by many clothing types. Therefore, we learn a semantic clothing prior from a large-scale clothed human scan dataset, which has varied subjects, poses and camera views to which the SMPL body is fitted~\cite{agora}. This prior encodes the likelihood of clothing types given a vertex on the SMPL body model, which gives the correspondence between segmentation labels and the SMPL body surface. Then, we use this prior to calculate a loss between the SMPL body and observed clothing labels in images. To achieve this we introduce \textbf{\textit{D}}ifferentiable \textbf{\textit{S}}emantic \textbf{\textit{R}}endering (\textbf{\textit{\modelname}}), a novel loss that supervises the training of 3D body regression with clothing semantics using weak supervision~\cite{graphonomy}.

\begin{figure}[t]
    \includegraphics[width=0.5\textwidth]{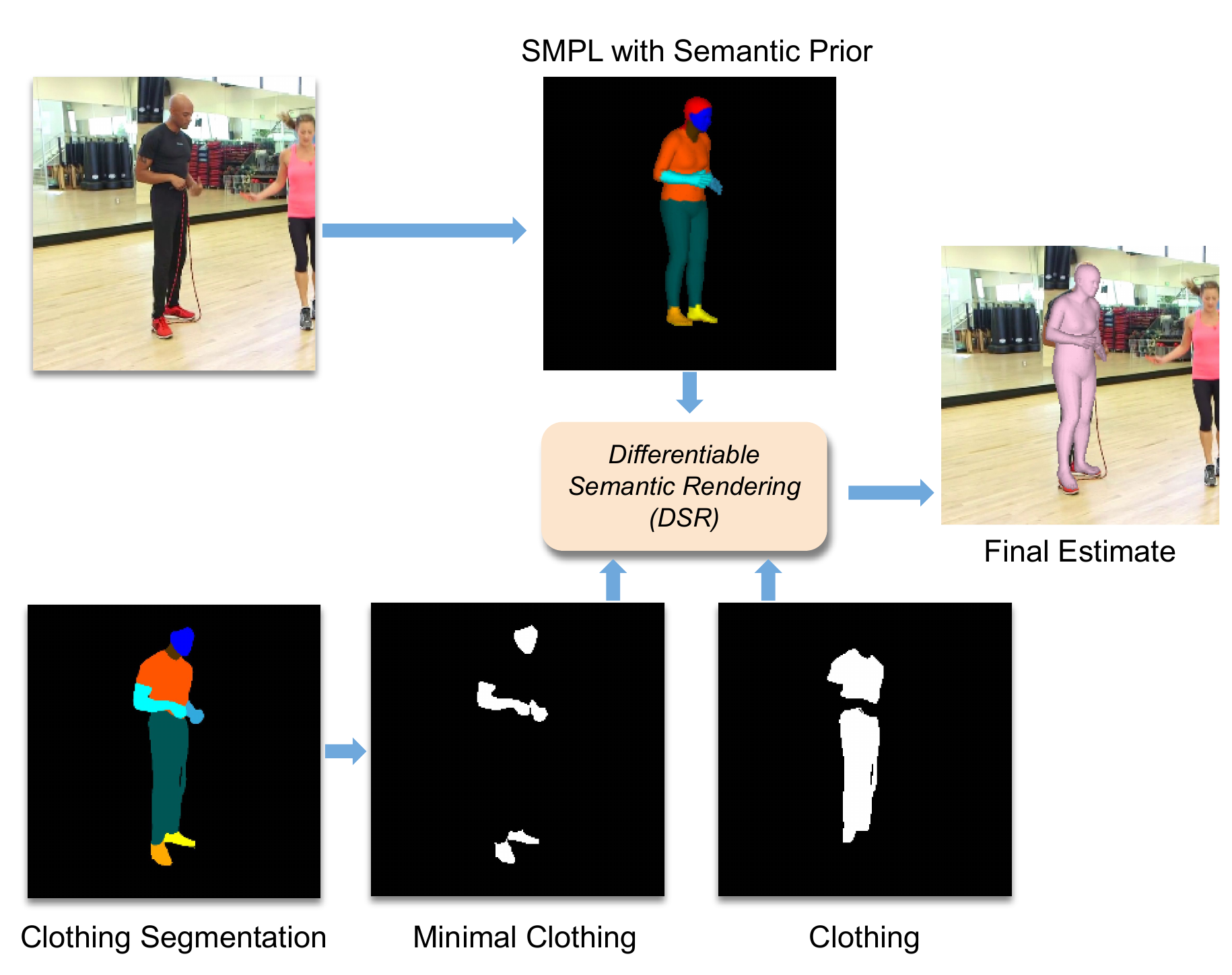}
    \caption{\textbf{DSR Idea} - For more accurate human body estimation, we supervise 3D body regression training with clothed and minimal-clothed regions differently using our novel \emph{DSR} loss and our learned semantic prior. The semantic prior represents a distribution over possible clothing labels for each vertex. For easier illustration, we depict the most likely labels per-vertex here.}
    \label{fig:idea}
\end{figure}

Our novel loss has two components: \textit{\modelname-C} for supervising the clothing region and \textit{\modelname-MC} for the minimally-clothed region. A high-level illustration of our idea is shown in Fig.~\ref{fig:idea}. While the former ensures that the rendered SMPL mesh stays inside the observed clothing label, the latter tries to tightly match the rendered SMPL mesh to the 2D minimal-clothing mask. The loss between the rendered output and the target mask is back-propagated using a differentiable renderer.
Specifically, for the ~\textit{\modelname-MC} term, we apply pixel-level supervision for tight-fitting with the minimal-clothing regions, while for \textit{\modelname-C}, we minimize the negative log probability of a SMPL semantic part label being inside the respective segmentation mask. For example, there will be a high penalty if the rendered vertices with a high probability of being ``shirt" fall in the ``pants" segmentation pixels. To ensure that our method is fast and differentiable, we render the semantic class probabilities computed from 3D scans as textures of the SMPL mesh. 

While training, DSR can be used as an additional loss in any neural network-based human body estimator that predicts SMPL parameters. 
First, we examine the effect of our approach over a baseline full-body mask supervision and 3D joint only supervision which verifies our hypothesis about the value of clothing semantics.
Then, we perform extensive comparisons and show that \modelname outperforms previous state-of-the-art methods as shown in Fig.~\ref{fig:teaser}. In summary:

\begin{enumerate}[noitemsep, topsep=1pt]
    \item We explore the importance of clothing semantics for 3D human body estimation by introducing a novel differentiable semantic rendering loss that distinguishes between clothed and minimally-clothed regions.
    \item We estimate a semantic clothing prior for SMPL from 3D scans of clothed people for our method which can be used also for other cases when a vertex clothing probability for a 3D SMPL body is required.
    \item We outperform all state-of-the-art methods on 3DPW and Human3.6M and obtain on par results on MPI-INF-3DHP, suggesting the value of using human parsing and semantics for more accurate human body estimation.
\end{enumerate}
\section{Related Work}
Estimating human pose and shape is a vastly growing field using different sources of supervision and input (image, video, keypoints, etc.). Here, we focus on different works that estimate 3D human pose and shape from an RGB image. We refer to recent surveys~\cite{Chen2020MonocularHP,SARAFIANOS20161} for more details.
\subsection{Image cues and 2D/3D joints}
Towards estimating  3D human pose and shape, initial attempts focus on estimating the coordinates of 3D joints or heatmaps~\cite{Hogg1983ModelbasedVA,Li20143DHP, Pavlakos2017CoarsetoFineVP, Sigal2004TrackingLP,SimoSerra2013AJM, Stoll2011FastAM, Sun2017CompositionalHP} from images using geometric assumptions for the human body and 3D training data. However, those approaches require 3D ground-truth data, which are limited in terms of pose variation, quantity and background, and lack generalization to in-the-wild images. The vast progress in 2D pose detection~\cite{Cao2021OpenPoseRM,Newell2016StackedHN,Sun2019DeepHR, Xiao2018SimpleBF}, along with the introduction of parametric body models of pose and shape~\cite{Anguelov2005SCAPESC,smpl} lead to significant progress and high-quality in-the-wild 3D humans. In~\cite{smplify} the authors use 2D keypoints to obtain SMPL parameters with an optimization-based approach, while this process improves via human annotations on predicted fits~\cite{Lassner2017UniteTP}. Martinez et al.~\cite{Martinez2017ASY} show that lifting the predictions of a 2D keypoint detector provides a reasonable baseline for the 3D pose. Pavlakos et al.~\cite{Pavlakos2018OrdinalDS} use additional ordinal depth annotations for weak 3D supervision. Kolotouros et al.~\cite{Kolotouros2019ConvolutionalMR} regress vertex locations using a sub-sampled SMPL mesh and Graph-CNNs. Xiang et al.~\cite{Xiang2019MonocularTC} extract joint confidence maps and 3D orientation information via CNNs and pair them with a deformable body model. Furthermore, in HMR~\cite{hmr}, a regressor from 2D joints to SMPL parameters is trained, using a discriminator with unpaired 3D data~\cite{Mahmood2019AMASSAO} to encourage plausible poses. Along these lines, some recent approaches using video as input, have applied similar methods to predict temporal kinematics of 3D bodies~\cite{Kanazawa2019Learning3H} and estimate the body using temporal features and a motion discriminator~\cite{Kocabas2020VIBEVI}. Another approach~\cite{Yu2019HumanMR}, uses a disentanglement of the skeleton from the 3D human mesh paired with a self-attention network to ensure temporal coherence. SPIN~\cite{spin}, revisits optimizations methods in collaboration with neural networks as it uses a network~\cite{hmr} that provides an initial estimate to the optimization process (SMPLify). Moreover, a regressor-based alternative suggests the use of the 3D neural regressor as a pose prior~\cite{eft}. Although such methods produce promising results, they typically estimate average body shapes, are not robust to occlusion, and produce poses that are only approximate. Without 3D training data the problem is hard.

\begin{figure*}[]
    \includegraphics[width=\textwidth]{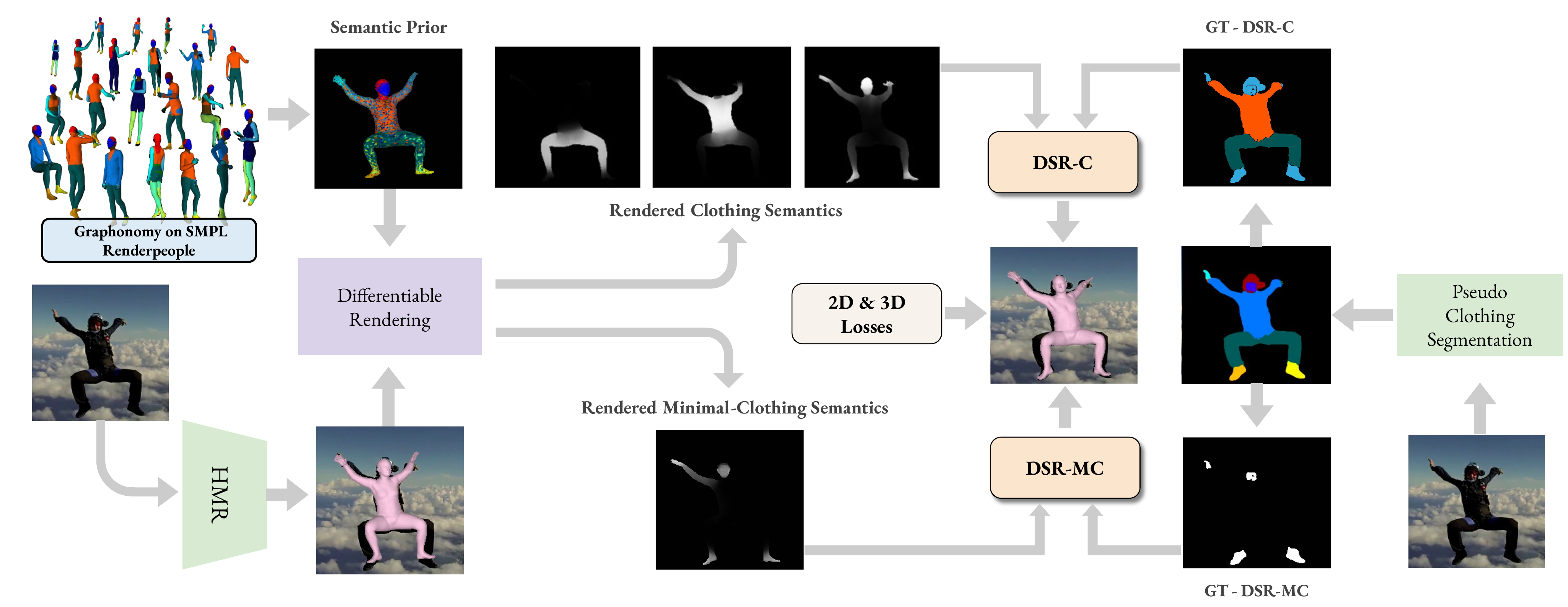}
    \caption{\textbf{Illustration of {\modelname}} - SMPL is rendered with the \emph{semantic prior} learned from RenderPeople scans. The two novel loss terms are calculated based on different semantic regions of the clothed person. \modelname-MC tightly fits the minimal-clothed region, while \modelname-C ensures that the rendered body lies within the clothing boundaries.}
    \label{fig:method}
\end{figure*}

\subsection{Image alignment and pixel-level supervision}
Concurrently, there is a line of research that uses additional constraints, in addition to image features and 2D/3D joints, to better align the body with the image such as dense body landmarks, silhouettes, body part segmentation or pixel aligned implicit functions.
Initial seminal lines of work, use a few keypoints along with the SCAPE body model and optimize 3D body shape with silhouettes and smooth shading~\cite{Guan2009EstimatingHS}. Along these lines, Balan et al.~\cite{balan_eccv} propose a distance function for the connected silhouette to ensure the rendered 3D model falls inside the mask. Later work uses 2D keypoints, background segmentation, and SMPL to extract 3D bodies from images~\cite{Tung2017SelfsupervisedLO}, similar to \cite{pavlakos2018} who use silhouettes for supervision. Even silhouettes, although they provide supervision when keypoints fail, are often ambiguous in the case of self-occlusion. Towards a detailed alignment of the 3D human body surface and pixels the authors in~\cite{Gler2018DensePoseDH} introduce a dataset with image-to-surface correspondence from MS-COCO~\cite{coco} and a variant of Mask-RCNN that regresses UV coordinates from images. Part-segmentation masks and IUV are also used in~\cite{nbf} and~\cite{densecor}, respectively, as dense supervision. A continuous UV map of SMPL for direct pixel correspondence of the image and the 3D mesh is introduced in~\cite{densecor}. In a similar approach~\cite{denserac}, exploits IUV maps as a proxy representation. It estimates SMPL parameters by minimising dense body landmarks and human part masks and also by using motion discriminator. While a large majority of the aforementioned work leverages a parametric 3D body model, there is some recent work that uses voxel representations along with 2D pose and part segmentation supervision~\cite{Varol2018BodyNetVI} or employs implicit functions with surface reconstruction techniques to reconstruct clothed humans. Although these approaches output fine-grain details, they are unable to capture the shape under clothing and are prone to occlusion~\cite{He2020GeoPIFuGA, Saito2019PIFuPI, saito2020pifuhd}. An interesting approach is proposed in~\cite{Zhang2020ObjectOccludedHS} where a partial UV map of the person is used and the human pose estimation is formulated as an image inpainting problem. Another work that explores scene semantics~\cite{Rafi2015ASO}, predicts the label of an occluding object and employs this information to detect invisible joints. Finally, Zanfir et al.~\cite{Zanfir2020WeaklyS3} represent the body with a normalizing flows-based latent space and use body part segmentation supervision to estimate 3D human body pose from videos and images, unifying different previous approaches. Clothing segmentation is used in~\cite{Xiang2020MonoClothCapTT} for clothing deformation to penalize the vertex offset of the clothed body if the rendered vertex falls outside the clothing boundary.

Most of these approaches are based on joints, silhouettes and part-segmentation masks using approximate supervision for the pose of a person in clothing. 
We claim that there is more that an image can tell us about  human pose. Our key insight is that clothing for different parts of the body conveys important information for detailed fitting. We employ an off-the-shelf 2D semantic segmentation method~\cite{graphonomy} and a semantic clothing prior to apply these labels in 3D. Given those, we supervise clothed and minimal-clothed regions separately, yielding more aligned fits of 3D humans.
\section{Method}

{\modelname} uses high-level semantic information for more accurate pose and shape estimation using two additional loss terms {\modelname}-MC and {\modelname}-C, as shown in Figure~\ref{fig:method}. {\modelname} takes an image $I$ as input which passes through a CNN. Then, the image features $\Phi(I)$ are fed to an iterative regressor, similar to HMR~\cite{hmr}, to estimate the parameters of SMPL body model, $\hat{\Theta}$. Given the rendered SMPL mesh, we apply our novel \modelname-MC and \modelname-C losses in addition to standard loss terms used in EFT~\cite{eft}. 
SMPL is a parametric body model that represents body pose and shape by $\Theta = [ \theta \in \mathbb{R}^{72}, \beta \in \mathbb{R}^{10} ] $. The pose parameters $\theta$ include the global rotation and rotations of $23$ body joints in axis-angle format and the shape parameters $\beta$ consist of the first 10 coefficients of a PCA shape space. SMPL model is a differentiable function $\mathcal{M}(\theta, \beta) \in \mathbb{R}^{6890 \times 3}$ that outputs a 3D mesh according to the pose and shape parameters.

\textbf{Clothing Semantic Information.} Ground-truth clothing segmentations are  expensive to obtain for in-the-wild datasets, which limits the scalability of such an approach. Hence, to analyze the importance of clothing semantics for human pose and shape estimation, we employ an off-the-shelf segmentation model to generate pseudo ground-truth clothing semantics.
Graphonomy~\cite{graphonomy} is a state-of-the-art clothing segmentation model that uses inter and intra graph transfer learning for unifying different clothing datasets and produces $20$ clothing labels and body part segmentations. As \modelname-MC reasons about the minimal-clothing region, we use a binary mask comprised of 5 labels - \emph{LeftArm, RightArm, LeftShoe, RightShoe and Face} from Graphonomy as ground-truth (whenever available). For \modelname-C, we use 4 labels - \emph{UpperClothes, LowerClothes, MinimalClothing and Background}. 
We run the \emph{Universal Model} of Graphonomy on all the datasets to generate pseudo-truth clothing segmentations. For more details on the generation of pseudo-ground truth, cleaning of obtained masks and mapping of graphonomy labels for \modelname-C and \modelname-MC, please refer to the Sup.~Mat.

\textbf{Semantic Prior for SMPL.} To use the semantic information obtained from Graphonomy as pseudo ground-truth training labels, we need a semantic prior of clothing for SMPL 3D bodies. To achieve this, we use thousands of scans from Renderpeople~\cite{renderpeople} with varied clothing, subject, pose and $10$ camera views for which we have ground-truth SMPL fits from AGORA~\cite{agora}. We run the universal model of Graphonomy on the rendered images of the scan with $20$ clothing and body part segmentation labels. Next, we use the ground-truth SMPL mesh to compute the visible face triangles given the mesh and camera parameters. Then, each visible triangular face is assigned the corresponding segmentation label. We repeat this process for all the available scans. We compute the probability of each vertex being a particular label out of the $20$ labels from Graphonomy. This probabilistic label for each vertex is referred to as \emph{semantic prior}. For more details refer to Sup.~Mat. 

\textbf{Differentiable Semantic Rendering.} We use SoftRas~\cite{softras} as the differentiable renderer to supervise the estimation of the 3D parametric model using semantic information. It uses a differentiable aggregation process for rendering, which fuses the probabilistic contributions of all mesh triangles with respect to rendered pixels. The semantic prior obtained from AGORA~\cite{agora} is used as a texture. Specifically, for each semantic label, we render the probability of that label for each visible vertex. Once the semantic probability is rendered as images by SoftRas, the loss is computed on the 2D image output by comparing with the semantic image segmentation and this is backpropagated to change the vertices, in turn, changing the network to give more accurate SMPL parameters. 

\textbf{Standard Losses.} As we use the EFT~\cite{eft} data for training, we use the standard supervision loss $\mathcal{L}_{SD}$ similar to EFT which is defined as:
\begin{equation}
    \begin{gathered}
 \mathcal{L}_{2D}(\Pi(\mathcal{M}(\hat{\Theta})), j) + \mathcal{L}_{3D}(\mathcal{M}(\hat{\Theta}), J) + \mathcal{L}_{\Theta}(\Theta, \hat{\Theta})
    \end{gathered}
\end{equation}
where, $\hat{\Theta}$ are the estimated SMPL parameters, $\mathcal{L}_{2D}$ is the joint reprojection loss, $\mathcal{L}_{3D}$ and $\mathcal{L}_{\Theta}$ are losses on 3D joints and SMPL parameters, respectively. Ground truth 2D joints are represented by $j$, 3D joints by $J$, SMPL parameters by $\Theta$ and the camera projection function by $\Pi$.

\textbf{DSR - Minimal-Clothing.} For minimal-clothing, we choose five labels from Graphonomy namely, \emph{LeftArm, RightArm, LeftShoe, RightShoe and Face}, which often appear similar in shape to the rendered SMPL body; i.e.~look roughly ``naked."
For a particular image, we take the clothing segmentation mask given by Graphonomy and create a binary mask $G$ comprising of the valid labels for that image from the available five labels. This forms the ground-truth for \modelname-MC denoted by \emph{GT - DSR-MC} in Fig.~\ref{fig:method}. We render the probability distribution of vertex labels for SMPL precomputed from RenderPeople as textures; these are shown as \textit{Rendered Semantics} in Fig.~\ref{fig:method}. We only take the probability distribution of vertices that are visible and set the others as zero. Thus, we define the \modelname-MC loss to tightly match the corresponding rendered minimal-clothing region of SMPL to the available semantic binary mask as shown in Fig.~\ref{fig:method} (bottom).

We study two variants of the loss for \modelname-MC: soft-DistM and soft-IOU. Soft-DistM is inspired by the DistM loss of Naked Truth~\cite{balan_eccv} which was originally proposed for estimating body shape under-clothing. Since we render the semantic probability instead of silhouettes, we call it soft-DistM. It is a distance measure function that takes the rendered image $R$ and target binary Graphonomy mask $G$ and is defined as: 
\begin{equation}
    \mathcal{L}_{MC-sDistM} = \sum_{i,j} (R_{i,j} \cdot d_{i,j}(G)) / {(\sum_{i,j} R_{i,j})}^{3/2}
\end{equation}
where $R_{i,j}$ are the pixels inside rendered human body and $d_{i,j}$ is a distance function which is zero if pixel $(i,j)$ is inside $G$. For points outside, it is defined as the Euclidean distance to the closest point on the boundary of $G$. Soft-DistM can pull the output inside the target because of the sharp difference in penalization between pixels inside the mask and pixels outside. Given a good initial estimate, the Soft-DistM loss ignores spurious and scattered labels outside the region of interest because the loss is high for pixels far away. This is particularly helpful, when using an off-the-shelf segmentation model without instance segmentation, which can give the wrong output for hard examples.

However, soft-DistM cannot fully ensure that the rendered output exactly matches the target as it gives the same penalty for outputs with different percentages of overlap when it is inside the boundary. Hence, we studied soft-IoU, which ensures tight fitting and is calculated as:
\begin{equation}
    \mathcal{L}_{MC-sIOU}= \frac{1}{N} \frac{ \sum_{(i,j)} P_{i,j} \cdot {G}_{i,j}}{\sum_{(i,j)} P_{i,j} + {G}_{i,j} - P_{i,j} \cdot {G}_{i,j}}
\end{equation}
where $P_{i,j}$ is the rendered vertex probability at pixel $(i,j)$, $G_{i,j}$ is the graphonomy label for that pixel. Soft-IoU suffers from spurious and scattered labels outside the region of interest and also suffers from the lack of instance segmentation in the off-the-shelf model. However, we choose soft-IoU for the metric for \modelname-MC due to better quantitative results in the baseline experiments in Table~\ref{tab:baseline}.

\textbf{DSR - Clothing.} The rendered SMPL body mesh cannot exactly match all the target pixels for the clothing region. Hence, for a more accurate estimate of the 3D body model, we want to encourage the rendered SMPL mesh to stay inside the clothing mask. Previous methods~\cite{balan_eccv} define a distance function to deal with such scenarios. However, we have higher-level semantic information than a silhouette to better address this. We have additional boundaries other than the body outline to enforce that a particular semantic part of the SMPL mesh should fall inside the corresponding semantic part of the segmentation mask. Clothing segmentation provides additional boundaries, such as between the upper and lower body or between clothing and skin.

Specifically, we define four labels, \emph{UpperClothes, LowerClothes, MinimalClothing and Background}, shown as four color masks in Fig.~\ref{fig:method}~(top). We  introduce a~\emph{MinimalClothing} label for \modelname-C to avoid  confusion between the background and minimal-clothing region.  Without it, the \modelname-C loss would give the same penalty when the minimal-clothing region falls on the corresponding target region or the background. As the semantic prior learned from RenderPeople has $20$ probability labels per vertex, we add all the probabilities of upper body clothing labels for~\emph{Upperclothes}, lower body clothing labels for~\emph{LowerClothes} and body part segmentation labels for~\emph{MinimalClothing}. We define \modelname-C loss as the negative log-likelihood~(NLL) of the rendered probability distribution of each vertex belonging to one of the four labels. The rendered probability distribution is first sent through log softmax before applying NLL loss for numerical stability. So, $\mathcal{L}_{\modelname-C}$ is defined as 
\begin{equation}
    \mathcal{L}_{\modelname-C} = \sum_{i=1}^{W}\sum_{j=1}^{H} -log(y_{i,j})
\end{equation}
where $y_{i,j}$ is the probability output for the vertex at pixel $(i,j)$, $H$ is the height and $W$ is the width of the image. Hence, the total loss $\mathcal{L}_{total} = \mathcal{L}_{SD} + \mathcal{L}_{MC-sIOU} + \mathcal{L}_{\modelname-C} $.
\begin{figure}[t!]
    \centerline{\includegraphics[width=\linewidth]{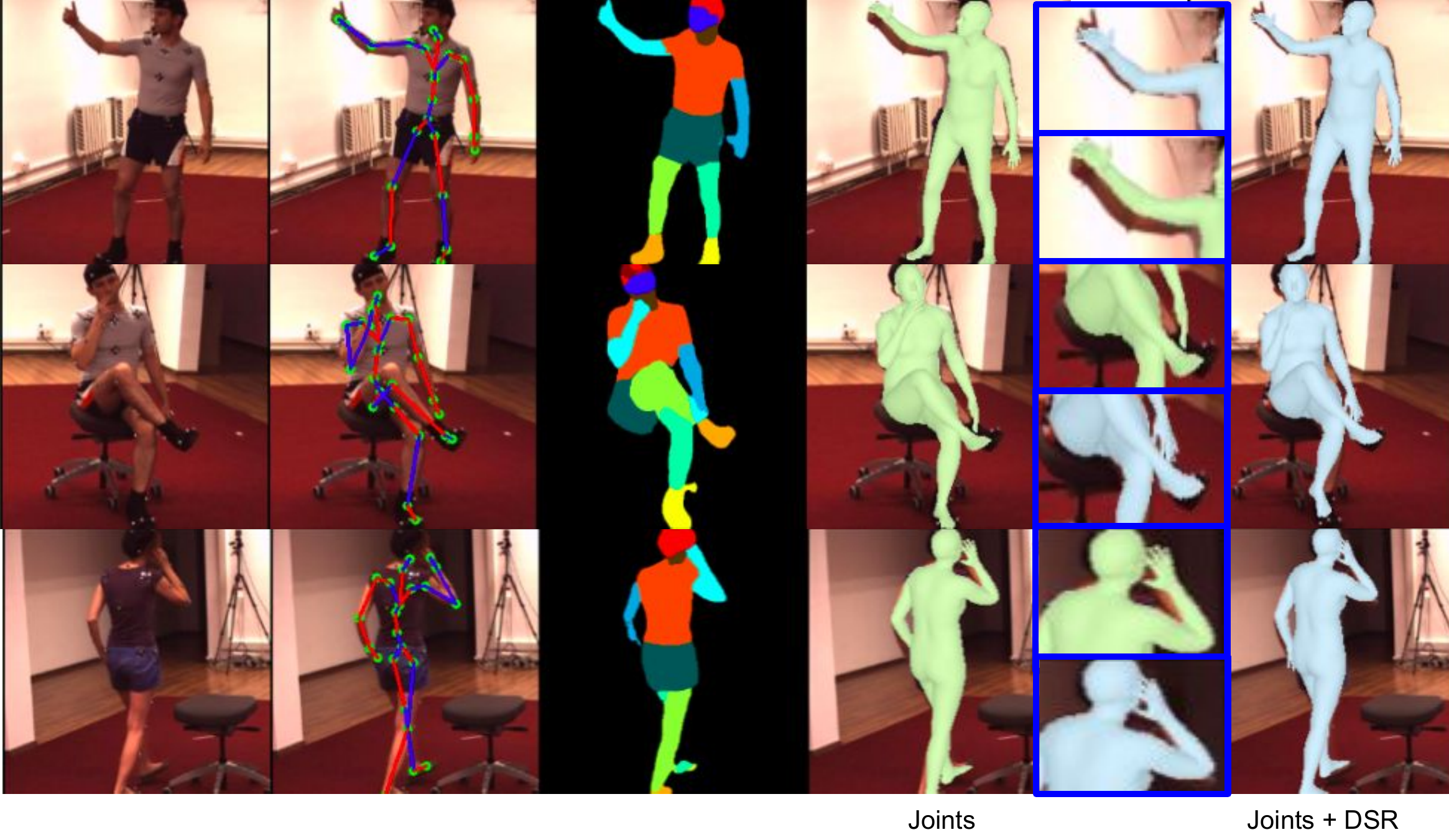}}
    \vspace{-0.1in}
    \caption{\textbf{Are 3D joints enough?} We over-fit a batch of H36M samples on ground-truth (GT) joints (green) and joints with DSR (blue). The weak supervision with semantic information improves accuracy.}
    \label{fig:overfitting_h36m}
\end{figure}

\section{Experimental Setup}

\textbf{Training Procedure.} Following EFT~\cite{eft}, we train a regressor similar to HMR \cite{hmr} with mixed 3D and 2D datasets. We use the pseudo-ground 3D annotations for 2D datasets from EFT. For 2D data, we only use COCO~\cite{coco} as including other in-the-wild datasets did not give a performance gain and for 3D datasets, we use Human3.6M~\cite{h36m_pami} and  MPI-INF-3DHP~\cite{Mehta2017Monocular3H}. We also use the 3DPW~\cite{vonMarcard2018} training set for fair comparisons and the same data ratio for mixed 2D and 3D datasets as EFT. For baseline and ablation experiments, we train only on COCO-EFT~\cite{eft}. For faster training, we initialize the network with SPIN pre-trained weights and use the same hyper-parameters as SPIN~\cite{spin} and train the model for 100K iterations.

\textbf{Evaluation Procedure.}~For state-of-the-art comparisons, we use 3DPW~\cite{Li20143DHP}, Human3.6M~\cite{h36m_pami} and MPI-INF-3DHP~\cite{Mehta2017Monocular3H}. As in prior work~\cite{eft, spin}, we use the gender information for ground truth meshes on 3DPW. We report results with and without 3DPW training on Procrustes-aligned mean per joint position error (PA-MPJPE), mean per joint position error (MPJPE) and Per Vertex Error (PVE). 

\textbf{Differentiable Semantic Rendering.}~We use SoftRas~\cite{softras} to render the probability distribution for \modelname-C and \modelname-MC. For SoftRas, we use a higher gamma value of $1.0 \times 10^{-1}$ to ensure the loss affects the occluded part of the body and a lower sigma value of $1.0 \times 10^{-5}$ to ensure the error does not significantly affect the spatial region. For more details, refer to SoftRas~\cite{softras}. We render the probability distribution of each triangle face as textures and compute the loss on the RGB channel of the rendered output. 
We render 5 images for each sample in a batched manner: 1 for \modelname-MC and 4 for \modelname-C. However, the loss is calculated per individual sample to avoid calculating for samples that do not have a valid segmentation mask. In such cases, the loss is set to zero. After using the heuristics to clean the mask, a valid label set is created for \modelname-C and \modelname-MC. The weighting parameters for both the components are set to $0.01$. As {\modelname} depends on weak supervision from off the shelf clothing segmentation model and hence not robust for hard examples, we enable the loss after 10K iterations into our training.
\begin{table}[t]
\centering
\resizebox{\linewidth}{!}{
     \begin{tabular}{l | c  c  c}
        \toprule
        Method &  PAMPJPE $\downarrow$ & MPJPE $\downarrow$ & PVE $\downarrow$\\
        \midrule 
        C-EFT & 58.5 & 101.0 & 119.3 \\
        \cellcolor{Gray}\hphantom{l}+ DSR-FB & \cellcolor{Gray}59.8 & \cellcolor{Gray}102.1 & \cellcolor{Gray}120.3 \\
        \hphantom{l}+ DSR-FB (s-DistM) & 58.0 & 100.2 & 117.8 \\
        \cellcolor{Gray}\hphantom{l}+ DSR-MC (s-DistM) & \cellcolor{Gray}58.2 & \cellcolor{Gray}100.6 & \cellcolor{Gray}118.5 \\
        \hphantom{l}+ DSR-MC (s-IoU) & 58.0 & 100.3 & 118.1 \\
        \cellcolor{Gray}\hphantom{l}+ DSR-C & \cellcolor{Gray}57.6 & \cellcolor{Gray}99.8 & \cellcolor{Gray}117.6 \\
        \hphantom{l}+ DSR-MVP & 58.1 & 100.3 & 117.8 \\
        \cellcolor{Gray}\hphantom{l}+ DSR-C + DSR-MC (Ours) & \cellcolor{Gray}\textbf{57.2} & \cellcolor{Gray}\textbf{99.2} & \cellcolor{Gray}\textbf{116.3} \\
        \bottomrule 
     \end{tabular}
   }
   \vspace{-0.05in}
\caption{\textbf{Baseline Comparisons for DSR on 3DPW.} C-EFT is the regressor trained with COCO-EFT and standard losses. DSR-FB is supervised with a full-body silhouette. DSR-MC is minimal-clothing, DSR-C is clothing and DSR-MVP is manual labelling of clothing and minimal-clothing.}
\label{tab:baseline}
\end{table}

\begin{table*}[]
    \centering
    \resizebox{\textwidth}{!}{%
        \begin{tabular}{ll|ccc|cc|cc}
        
            \toprule

            &  & \multicolumn{3}{c}{3DPW} & \multicolumn{2}{c}{Human3.6M} & \multicolumn{2}{c}{MPI-INF-3DHP}  \\
            \cmidrule(lr){3-5} \cmidrule(lr){6-7} \cmidrule(lr){8-9}
            & Models & PA-MPJPE $\downarrow$ & MPJPE $\downarrow$ & PVE $\downarrow$ & PA-MPJPE $\downarrow$ & MPJPE $\downarrow$ & PA-MPJPE $\downarrow$ & MPJPE $\downarrow$ \\
            
            \midrule

            & HMR~\cite{hmr} & 76.7 & 130.0 & - & 56.8 & 88 & 89.8 & 124.2 \\
            & \cellcolor{Gray}NBF~\cite{nbf} & \cellcolor{Gray}- & \cellcolor{Gray}- & \cellcolor{Gray}- & \cellcolor{Gray}59.9 & \cellcolor{Gray}- & \cellcolor{Gray}- & \cellcolor{Gray}-  \\
            & Pavlakos \etal\cite{pavlakos2018} & - & - & - & 75.9 & - & - & -  \\
            & \cellcolor{Gray}CMR~\cite{Kolotouros2019ConvolutionalMR} & \cellcolor{Gray}70.2 & \cellcolor{Gray}- & \cellcolor{Gray}- & \cellcolor{Gray}50.1 & \cellcolor{Gray}- & \cellcolor{Gray}- & \cellcolor{Gray}-  \\
            &  SPIN~\cite{spin} & \ 59.2 &  96.9 &  116.4  &  41.1 &  62.5 & 67.5 &  \textbf{105.2} \\
            & \cellcolor{Gray}EFT~\cite{eft} & \cellcolor{Gray}54.2 & \cellcolor{Gray}- & \cellcolor{Gray}- & \cellcolor{Gray}43.7 & \cellcolor{Gray}- & \cellcolor{Gray}68.0 & \cellcolor{Gray}- \\
            & Zanfir et. al~\cite{Zanfir2020WeaklyS3} (w/ 3DPW train) & 57.1 & 90.0 & - & - &  - & - & -  \\
            & \cellcolor{Gray}EFT~\cite{eft} (w/ 3DPW train) & \cellcolor{Gray}52.2 & \cellcolor{Gray}- & \cellcolor{Gray}- & \cellcolor{Gray}43.8 & \cellcolor{Gray}- & \cellcolor{Gray}67.0 & \cellcolor{Gray}- \\
            
            \midrule
            
            &  DSR &  54.1 &  91.7 &  105.8 &  \textbf{40.3} &  \textbf{60.9} &  \textbf{66.7} & 105.3 \\
            &  \cellcolor{Gray}DSR (w/ 3DPW train) &  \cellcolor{Gray}\textbf{51.7} &  \cellcolor{Gray}\textbf{85.7} &  \cellcolor{Gray}\textbf{99.5} &  \cellcolor{Gray}41.4 &  \cellcolor{Gray}62.0 &  \cellcolor{Gray}67.0 & \cellcolor{Gray}104.7 \\
            
            \bottomrule
            
        \end{tabular}%
    }
    \vspace{-0.05in}
    \caption{\textbf{Evaluation of state-of-the-art models on 3DPW, Human3.6M, and MPI-INF-3DHP datasets.} 
    DSR is our proposed model trained on monocular images similar to~\cite{spin, eft, hmr}. DSR outperforms all state-of-the-art models, including EFT~\cite{eft} on the challenging datasets. ``$-$" shows the results that are not available.}
    \label{tab:sota}
\end{table*}{}

\begin{table}[]
\centering
\resizebox{\linewidth}{!}{
     \begin{tabular}{l | c  c  c}
        \toprule
        Method &  PAMPJPE $\downarrow$ & MPJPE $\downarrow$ & PVE $\downarrow$\\
        \midrule 
        Standard Loss (SD) & 47.5 & 73.9 & 99.2 \\
        \cellcolor{Gray}SD + {\modelname} & \cellcolor{Gray}\textbf{45.1} & \cellcolor{Gray}\textbf{71.3} & \cellcolor{Gray}\textbf{96.6} \\
        \bottomrule 
     \end{tabular}
   }
\caption{\textbf{Potential of DSR.} We train and test on a subset of Human3.6M to evaluate the full potential of \modelname loss. SD refers to standard joint loss.}
\label{tab:h36m_sub}
\end{table}

\section{Results}

\subsection{Baseline Comparison and Ablation Studies}\label{sec:baseline}
We perform baseline experiments to (1) motivate the use of semantic rendering and (2) study how the different terms and design choices contribute to the final result as shown in Table~\ref{tab:baseline}.
As a baseline, we use an HMR~\cite{hmr} based regressor trained on EFT-COCO~\cite{eft} data and report results on 3DPW~(C-EFT). Then, we supervise the baseline with an additional full-body silhouette (DSR-FB) which is a per pixel binary classification loss guided by differentiable rendering. The results deteriorate as the rendered SMPL body does not match the full body. We further train DSR-FB with the Dist-M loss in contrast to per-pixel classification to ensure all body parts (irrespective of clothing) stay inside the silhouette. The result in Table~\ref{tab:baseline} shows that explicit supervision with clothing semantics (Ours) outperforms the naive cloth-agnostic approach. We study the importance of estimating clothing semantics from scans in contrast to manual vertex painting (MVP) of semantic labels as the former gives a distribution over possible clothing labels ($20$) for each vertex whereas the latter would give only $1$. To quantitatively verify the benefit of the probabilistic clothing semantic prior, we take the most likely label per vertex (Fig.~\ref{fig:idea}) as a proxy for MVP. Since we have 1 label per vertex, we use IoU instead of s-IoU. Table~\ref{tab:baseline} shows low performance of a fixed semantic prior (MVP) compared to a probabilistic one (Ours). We also study the individual contribution of \modelname-C and \modelname-MC on the overall performance and find that the clothing term helps more than the minimal-clothing term. One possible explanation could be that the off-the-shelf segmentation model is not robust for hands and feet hence causing less gain. Empirically, we observe that soft-IoU performs better than soft-DistM and hence use it as the metric for \modelname-MC for all subsequent experiments. Overall, the best accuracy is reached when both terms are used showing that supervising minimally-clothed and clothed regions differently helps improve 3D body estimation.

\begin{figure*}[t]
    \centerline{\includegraphics[width=\textwidth]{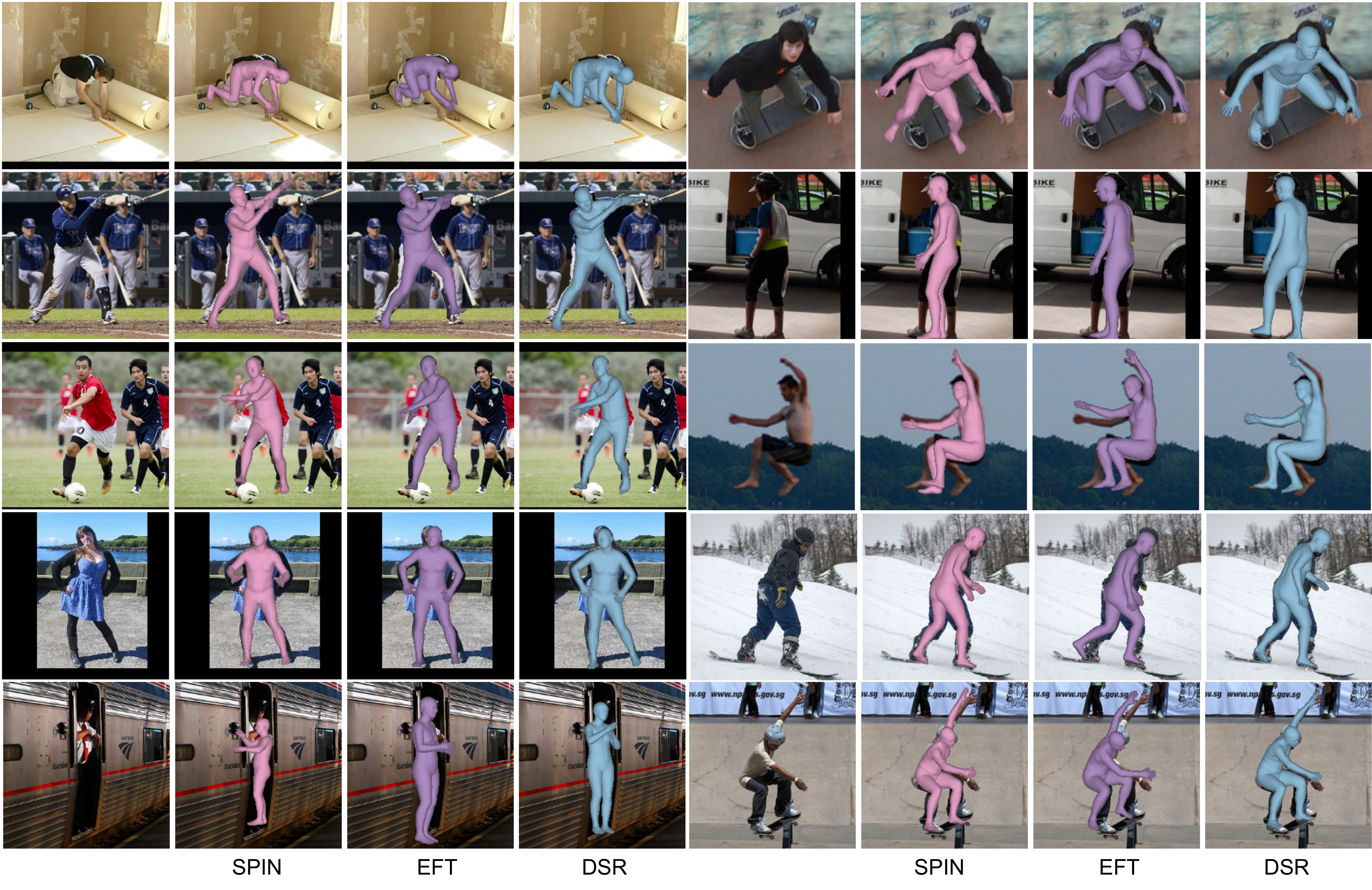}}
    \vspace{-0.15in}
    \caption{\textbf{Qualitative Results on COCO.} From left to right - Input image, SPIN~\cite{spin}, EFT~\cite{eft} and {\modelname}  results.}
    \label{fig:qual_result}
\end{figure*}

\subsection{State-of-the-art comparison}\label{sec:sota}
We compare our approach with state-of-the-art methods in Table~\ref{tab:sota}. We use two variants of our model, with and without the 3DPW training set, to be aligned with the training data of other methods. In 3DPW, an in-the-wild challenging 3D dataset, we outperform previous work when using 3DPW training data, while performing on par with EFT~\cite{eft} when they are not used. Moreover, we clearly improve accuracy on Human3.6M~\cite{h36m_pami}, a standard indoor benchmark, over state-of-the-art SPIN~\cite{spin} and EFT~\cite{eft} methods. We also report on par results in MPI-INF-3DHP~\cite{Andriluka20142DHP}. We perform significantly better than previous approaches that use ground-truth part-segmentation or silhouettes~\cite{pavlakos2018, nbf,Zanfir2020WeaklyS3} compared to our weak supervision. Overall, we consistently perform better than previous approaches across different datasets, both indoors and outdoors. In Fig.~\ref{fig:qual_result} we can see different comparisons of DSR with the previous state-of-the-art and observe that the estimated mesh is more aligned with image evidence. These observations validate our hypothesis that clothing semantics, even when used as weak supervision, provides additional information for estimating more accurate 3D bodies. 

\begin{table}[]
\centering
\resizebox{\linewidth}{!}{
     \begin{tabular}{l | c  c  c c c c }
        \toprule
        Method &  Ankle & Knee & Hip & Wrist & Elbow & Head \\
        \midrule 
        Standard Loss (SD) & 99.3 & 54.6 & 20.0 & 109.5 & 81.1  & 81.8 \\
        \cellcolor{Gray}SD + {\modelname} & \cellcolor{Gray}\textbf{96.1} & \cellcolor{Gray}\textbf{50.4} & \cellcolor{Gray}\textbf{19.5} & \cellcolor{Gray}\textbf{107.3} & \cellcolor{Gray}\textbf{79.3} & \cellcolor{Gray}\textbf{80.5} \\
        \bottomrule 
     \end{tabular}
   }
\caption{\textbf{Per joint error for Human3.6M subset.} SD refers to standard joint loss used in 3D body estimation.}
\label{tab:per_joint}
\end{table}

\subsection{Potential of \modelname}
To test the significance of high-level semantics on shape and pose estimation, we use an off-the-shelf segmentation model~\cite{graphonomy}. However, such models are not robust to in-the-wild examples. Because we use the output of the model as pseudo ground-truth for supervision, it is hard to determine the full potential of our approach. Hence, we experiment on the Human3.6M dataset to test the \modelname loss in a more controlled setting. Human3.6M is an indoor dataset with significantly less background complexity as compared to outdoor datasets. Hence, it is ideal for testing the limit of \modelname. We study two different cases.
First, we split the training set of Human3.6M, with SMPL parameters computed by MoSh~\cite{mosh}, into training and validation sets with S8 in the validation set. This is done to evaluate the per-vertex-error (PVE) using the MoSh ground truth SMPL parameters, thus, giving insight into the shape estimation efficacy of our method. As shown in Table.~\ref{tab:h36m_sub}, the performance gain with the \modelname loss is significantly higher compared to the standard joint loss. This emphasizes the importance of semantic information. We also analyze the per joint error to understand the source of a performance gain as shown in the Table.~\ref{tab:per_joint}. Using the \modelname, the maximum performance gains are from \emph{Ankle, Knee, Wrist} which are common failures in 3D pose estimation.
Second, we take a step further to examine whether ground truth 3D joints are enough for accurate and pixel aligned body estimation. To this end, we take a random batch of $64$ samples from Human3.6M and over-fit on only joints and joints with the \modelname loss for $100$ iterations with the same hyper-parameters used for other experiments. The qualitative results are depicted in Fig.~\ref{fig:overfitting_h36m}. As we can see, supervision with ground 3D joints cannot always reason about all the pixels. Using \modelname produces more pixel-aligned fits, especially for hands and feet.
\section{Conclusion}
While huge progress has been made in estimating 3D human bodies, we are still far from estimating highly accurate 3D humans. We hypothesize that clothing semantics is an under-explored feature that can benefit 3D body estimation methods. Therefore, we introduce a novel method to exploit clothing semantics as weak supervision. Namely, we: (1) Introduce a novel differentiable loss that supervises clothed and minimally-clothed regions differently to ensure that the body lies inside the clothes for the former while tightly fitting for the latter. (2) Learn a semantic clothing prior, i.e.~a probability distribution over clothing labels for SMPL vertices, to apply our method efficiently. (3) Thoroughly evaluate our approach qualitatively and quantitatively, outperforming the state-of-the-art. (4) Analyze our method's components and show that clothing semantics, even as weak supervision, is a valuable complementary cue to 3D joints that improves the estimation of 3D bodies. Our experiments show the importance of such semantics, providing new insight into 3D human body estimation.

{\modelname} uses clothing as weak supervision, which can be limited in complex scenes with multiple people and occlusion.  Our method can be easily extended to pipelines that account for multiple people in the scene~\cite{Zanfir2018Monocular3P}. In the future, we should explore methods that model 3D clothing semantics, build a better prior for SMPL bodies or incorporate additional constraints to disambiguate scene semantics. 

{\bf Acknowledgements:}
We thank Sergey Prokudin, Chun-Hao P. Huang, Vassilis Choutas, Priyanka Patel, Radek Danecek and Cornelia Kohler for their feedback. This work was supported by the German Federal Ministry of Education and Research (BMBF): Tübingen AI Center, FKZ: 01IS18039B. 
\noindent{\bf Disclosure:} \url{https://files.is.tue.mpg.de/black/CoI/ICCV2021.txt}

{\small
\bibliographystyle{ieee_fullname}
\bibliography{egbib}
}

\newpage
\appendix
\setcounter{figure}{0} \renewcommand{\thefigure}{\Roman{table}}
\setcounter{table}{0} \renewcommand{\thetable}{\Roman{table}}
\begin{strip}%
 \centering \Large
 \textbf{%
  Learning to Regress Bodies from Images using \\ Differentiable Semantic Rendering\\
 \vspace{0.5cm} \textit{Supplementary Material} \vspace{0.5cm}
 }\\
\end{strip}
\section{Clothing Semantic Information}
It is difficult to obtain ground-truth clothing segmentation masks for in-the-wild datasets. Hence, we use  Graphonomy~\cite{graphonomy}, which is an off-the-shelf human clothing segmentation model that provides reasonably reliable pseudo-ground truth. 

\begin{figure*}[t]
    \includegraphics[width=\textwidth]{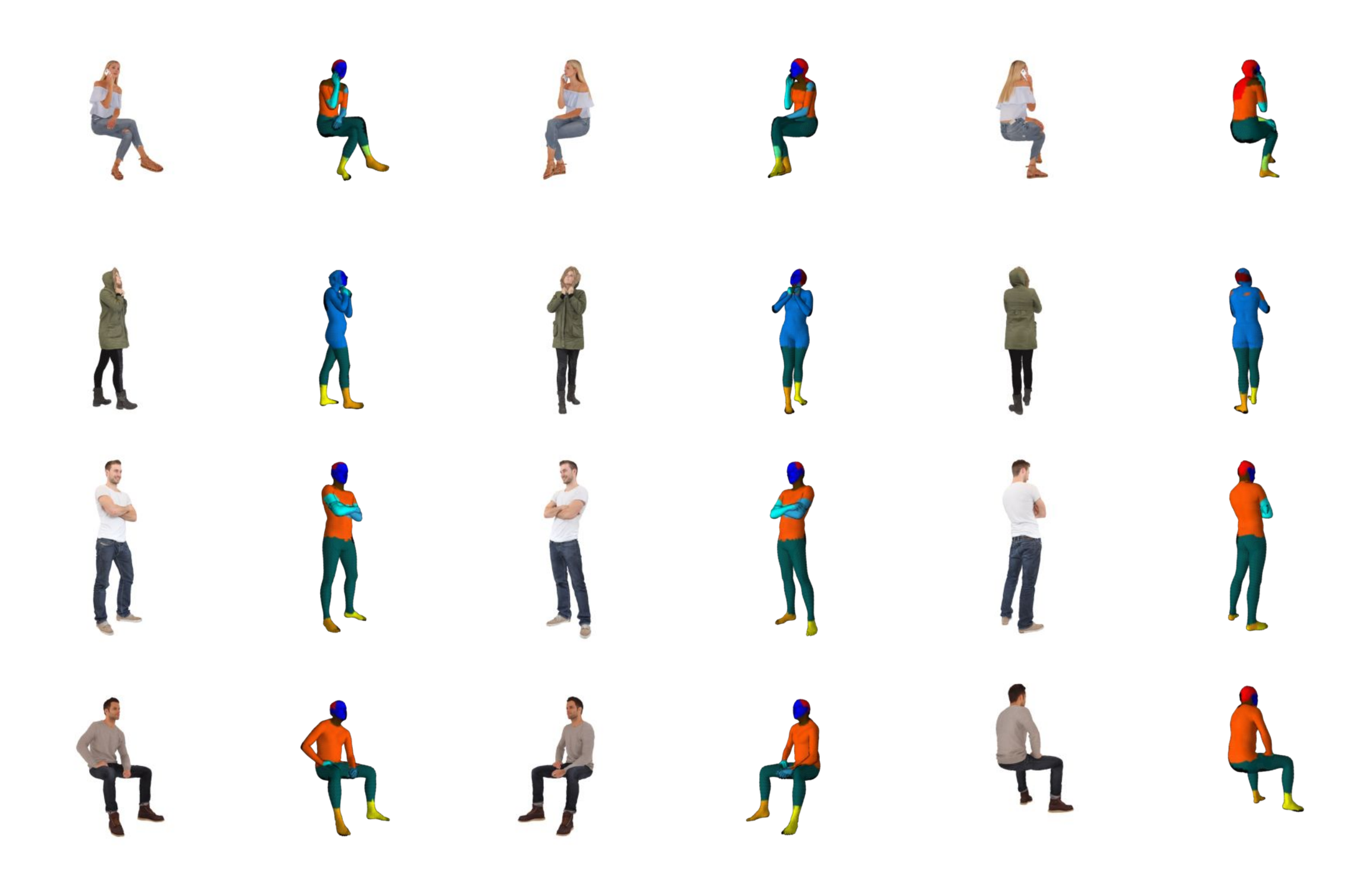}
    \caption{\textbf{Clothed Human Scans.} Examples of clothed human scans in different clothing, pose and camera views (\emph{Columns} 1,3,5) along with the corresponding SMPL bodies where each vertex is colored based on the output of the clothing segmentation model~\cite{graphonomy}~(\emph{Columns} 2,4,6) applied on the respective scan images. We only show $3$ camera views here. }
    \label{fig:sem_prior}
\end{figure*}
\subsection{Clothing Segmentation Masks} 
Graphonomy has three different models depending on the granularity of the segmentation mask and we choose the one with $20$ labels, also known as the \emph{Universal Model}. This model provides the best clothing segmentation performance compared to other Graphonomy variants. The different labels are:~\emph{Background, Hat, Hair, Glove, Sunglasses, UpperClothes, Dress, Coat, Socks, Pants, Jumpsuits, Scarf, Skirt, Face, LeftArm, RightArm, LeftLeg, RightLeg, LeftShoe} and \emph{RightShoe}. 

During inference, to get more accurate predictions -- as suggested in the original implementation -- we use $4$ different scaling factors for the input image -- $0.5, 0.75, 1.0, 1.5$ -- to account for different image resolutions. Then, we merge the outputs for different scaling factors using appropriate upsample and downsample functions (bilinear) to produce an output size the same as the original image. For images more than $1080 \times 1080$, we use a single scaling factor of $1.0$. We also flip the image horizontally and average the output predictions of the flipped image with the original one.

\subsection{Processing Pseudo Ground-Truth Masks} 
\label{subsec:proc}
The generated pseudo ground-truth cannot be directly used for supervising existing human body estimator networks because of incompatibility between Graphonomy's output and 3D pose regressor's training procedure~\cite{hmr}.

Graphonomy is not an instance segmentation model, which means it is hard to differentiate between people in the image. However, standard human body estimators~\cite{hmr, eft, spin} use a single person during training. To circumvent this problem, we use 2D keypoints to get a rough estimate of the region of the person in the image. Furthermore, we add/subtract an offset of $30$ pixels in both $x$ and $y$ direction according to the maximum/minimum keypoint location.

Due to occlusion or inaccuracies in the prediction, the spread of pixels for a particular label of Graphonomy may cover an extremely small part of the image. As \modelname-MC tries to tightly supervise the rendered SMPL body with the target binary mask, it is important to ensure the target masks are reliable. Hence, we remove labels that cover less than $60$ pixels from the predefined set of five labels \emph{(LeftArm, RightArm, LeftShoe, RightShoe, Face)}. 

There is a one to one mapping from the \modelname-MC labels to Graphonomy labels. The same is not true for \modelname-C as there are several clothing labels. 
Consequently, for \modelname-C, we define a coarse mapping as per Table~\ref{tab:label_mapping}.

\begin{table}[]
    \centering
    \resizebox{\linewidth}{!}{%
    \begin{tabular}{c|c}
       \toprule
        \modelname-C Labels & Graphonomy Labels \\
       \midrule
       \emph{Background} & Background \\
       \emph{LowerClothes} & Pants, Skirts \\
       \emph{UpperClothes} & Upperclothes, Dress, Coat, Jumpsuits \\
       \multirow{3}{*}{\emph{MinimalClothing}} & Hat, Hair, Glove, Sunglasses,\\ 
       & Socks, Scraf, Face, LeftArm, RightArm, \\
       & LeftLeg, RightLeg, LeftShoe, RightShoe \\
       \bottomrule
    \end{tabular}%
    }
    \caption{Mapping of \modelname-C labels to Graphonomy labels. }
    \label{tab:label_mapping}
\end{table}

\begin{figure}[t]
    \includegraphics[width=\linewidth]{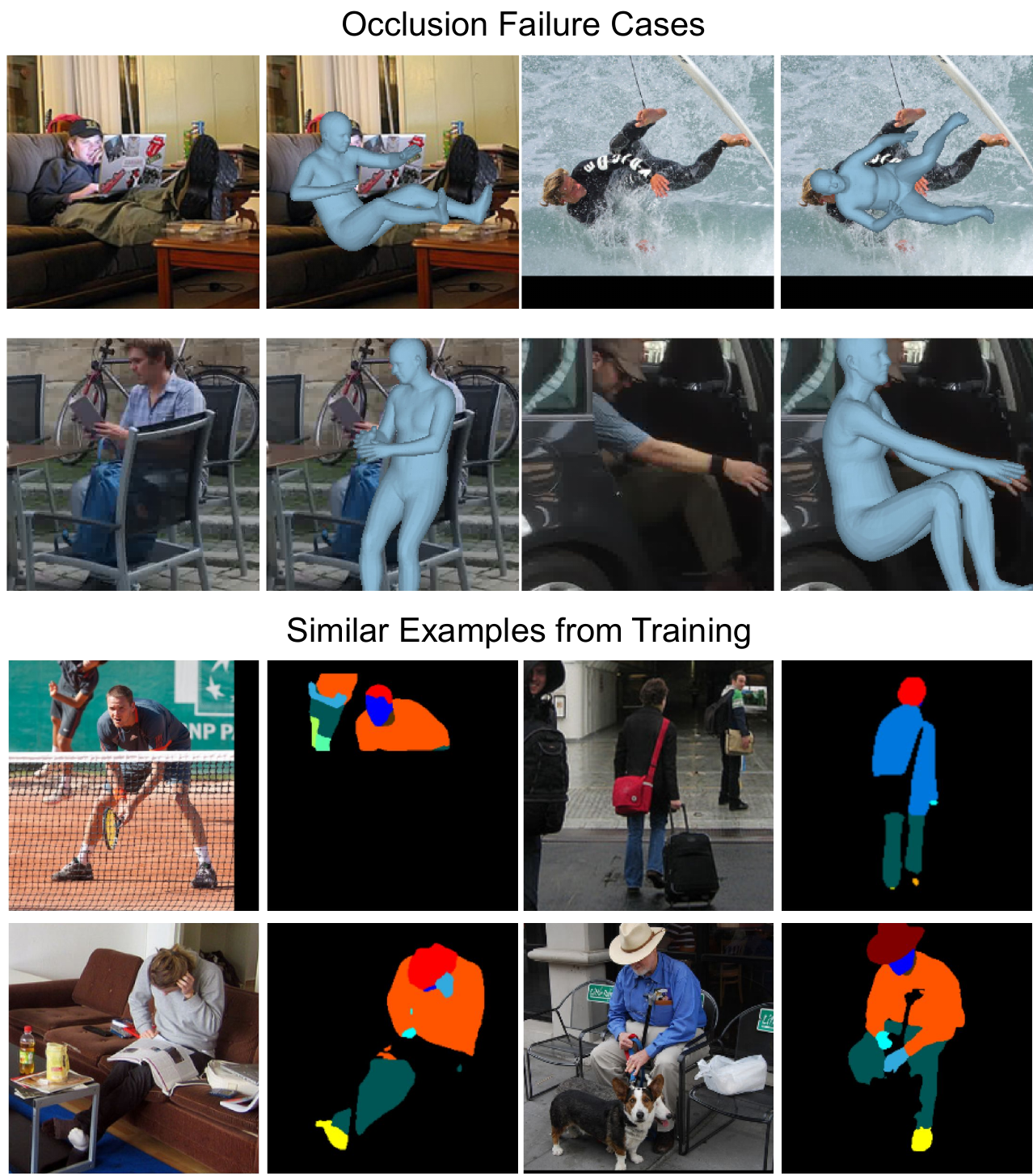}
    \caption{\textbf{Occlusion Failure Analysis} Qualitative failure results in case of occlusion. We show outputs from COCO and 3DPW in~\textit{Rows 1-2} respectively.  \textit{Rows 3-4:} Similar occlusion cases present in the training samples.}
    \label{fig:failure_occ}
\end{figure}

\section{Semantic Prior for SMPL}
To supervise the human body regressor network with semantic information, we need a term that captures the {\em a priori} probability that describes what parts of the SMPL body correspond to a particular semantic label. To this end, we use $2500$ clothed human scans from the AGORA dataset~\cite{agora} with varied clothing, pose and identity. 
AGORA contains clothed 3D people with ground truth SMPL-X bodies fit to the scans.
We convert SMPL-X fits to SMPL. For each scan, we render it from $10$ different camera views to cover different angles and generate scan images.
We run Graphonomy on each of these images to obtain $10$ 2D clothing segmentation images for each scan.
An illustration of the output from this process is depicted in Fig.~\ref{fig:sem_prior}.
We also render the fitted SMPL model with the known camera parameters to obtain the correspondences between the vertices of the SMPL body and the pixels in the image.

Given this training data, we can very simply compute the prior probability of a SMPL vertex having one of the $20$ Graphonomy labels.
We estimate this by calculating the occurrences of a particular label being present at the vertex divided by the total occurrences of other labels--excluding the \emph{Background} label.
Finally, this gives us the prior per-vertex probability that a SMPL vertex has given a Graphonomy label.
We also assign a small probability of a vertex being assigned the background label; this increases robustness to occlusion.
As an additional step, we use the SMPL body part segmentation to clean the semantic prior. Graphonomy gives incorrect predictions for some clothed body scan images and this will affect downstream tasks. 
Hence, if the semantic label probability of a ``leg'' vertex (denoted by SMPL part segmentation) has a higher probability of being hand, we set it to zero. 
This approach helps to avoid obvious failures when Graphonomy produces incorrect predictions.
Note that a more sophisticated prior model could also capture spatial correlations of clothing but we did not find this necessary.

\begin{figure}[t]
    \includegraphics[width=\linewidth]{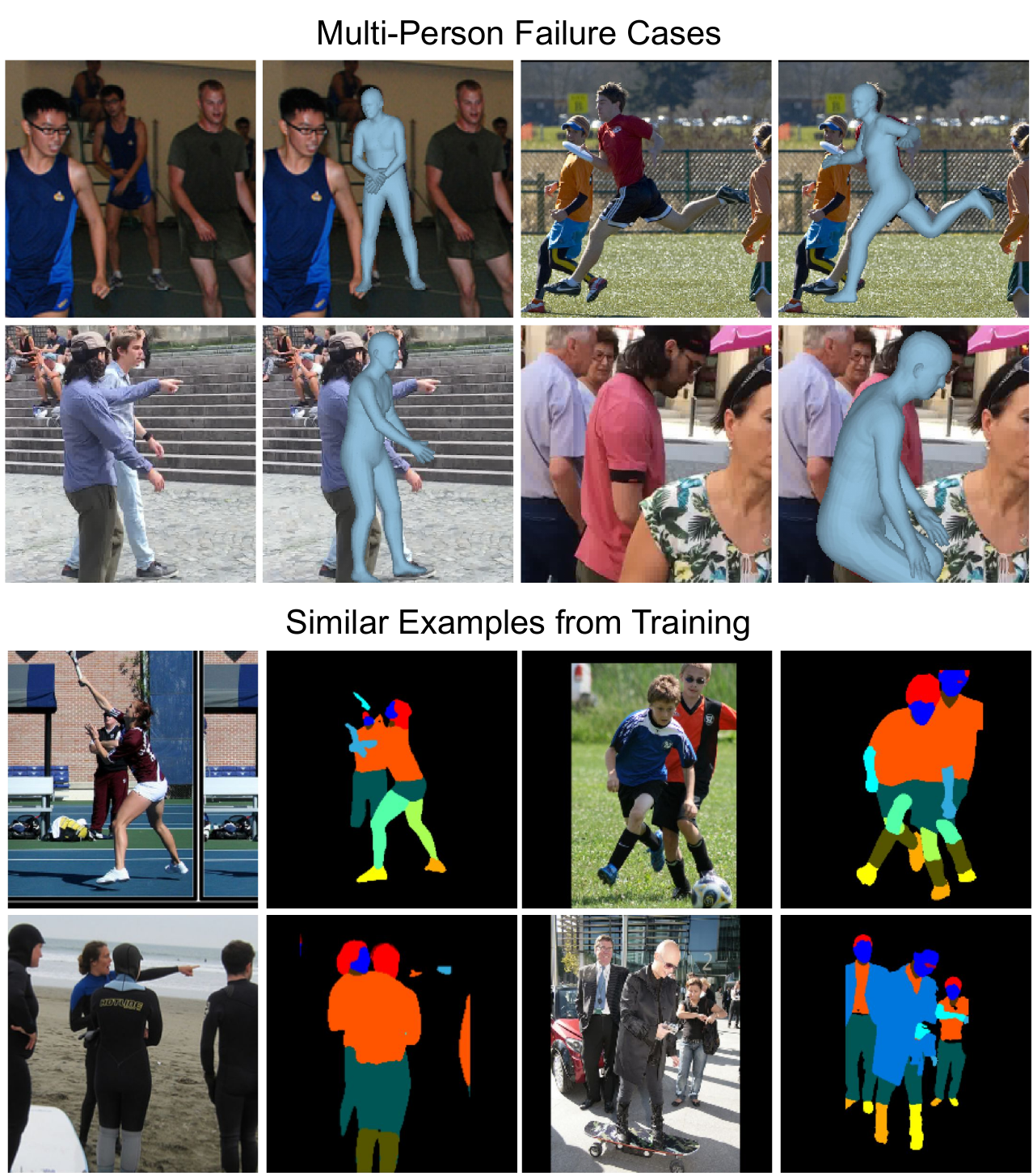}
    \caption{\textbf{Multi-Person Failure Analysis} Qualitative failure results in case multiple people are present. We show outputs from COCO and 3DPW in~\textit{Rows 1-2} respectively. \textit{Rows 3-4:} Similar multi-person failure cases present in the training samples.}
    \label{fig:failure_multiperson}
\end{figure}

\begin{figure*}[th]
    \includegraphics[width=\textwidth, height=0.95\textheight]{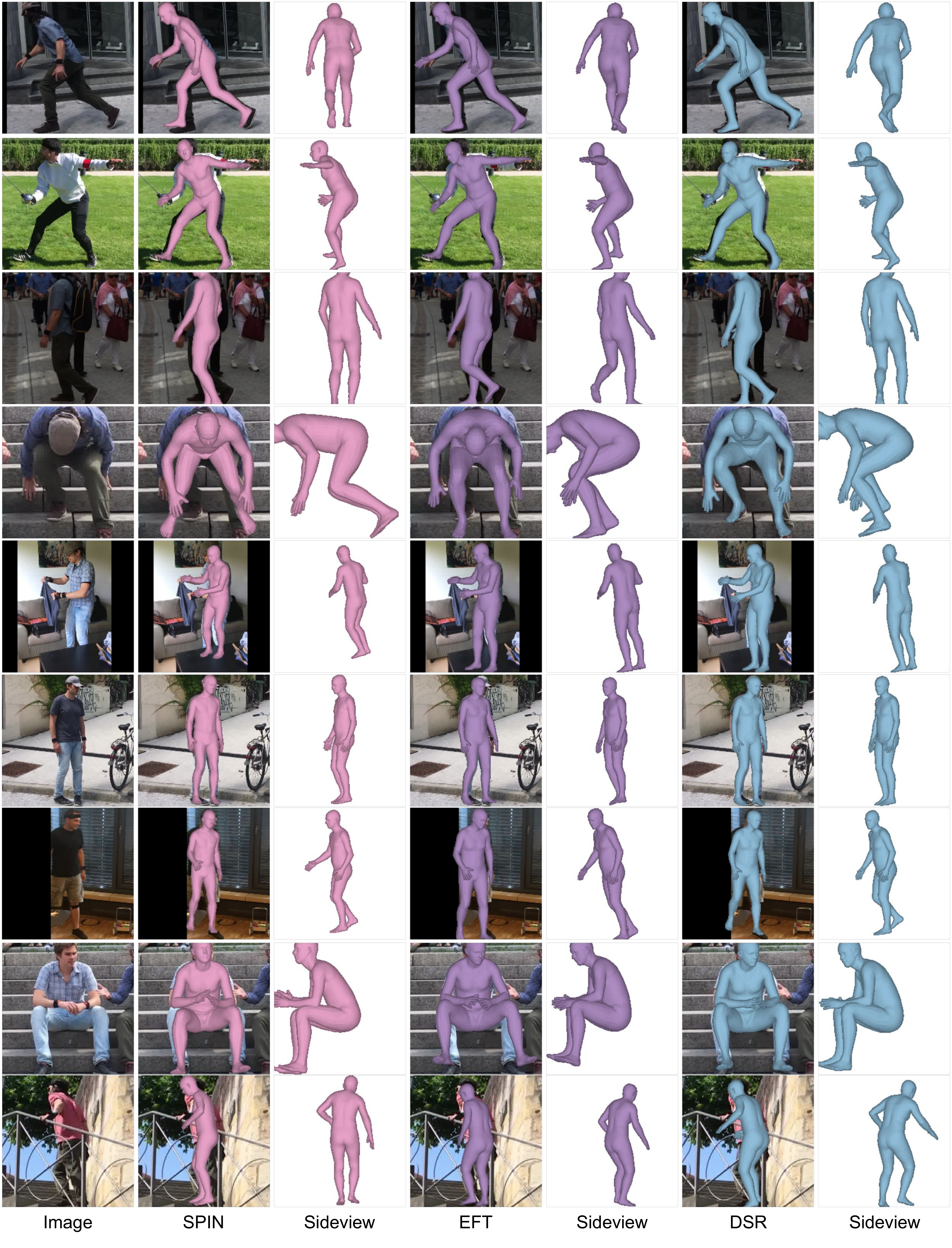}
    \caption{\textbf{Additional Qualitative Results of 3DPW.} From left to right - Input image, SPIN~\cite{spin}, SPIN Sideview, EFT~\cite{eft}, EFT Sideview, {\modelname} and \modelname Sideview  results}
    \label{fig:additional_qual_3dpw}
\end{figure*}

\begin{figure*}[th]
    \includegraphics[width=\textwidth, height=0.95\textheight]{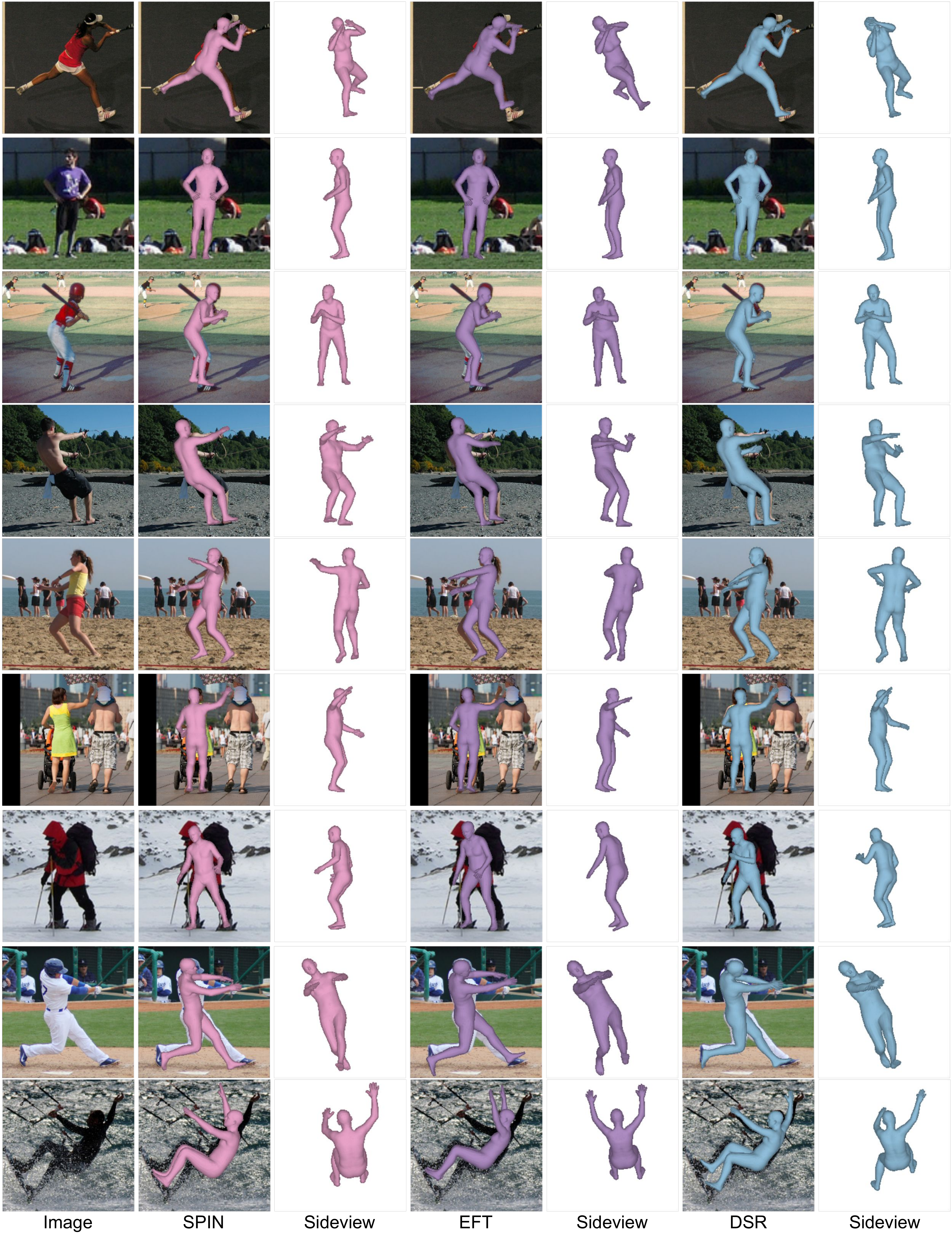}
    \caption{\textbf{Additional Qualitative Results of COCO.} From left to right - Input image, SPIN~\cite{spin}, SPIN Sideview, EFT~\cite{eft}, EFT Sideview, {\modelname} and \modelname Sideview  results}
    \label{fig:additional_qual_coco}
\end{figure*}

\section{Failure Case Analysis}
We qualitatively analyse the failure cases using our method and broadly categorise them into two types: occlusion failures as shown in Fig.~\ref{fig:failure_occ} and multi-person failures as shown in Fig.~\ref{fig:failure_multiperson}. Note that these are also cases where standard 3D pose estimation methods commonly fail.

First, we observe failures in case of either self-occlusion or scene occlusion producing unreasonable pose. Hence, we tried to analyse the training samples with occlusion. As we can see in Fig.~\ref{fig:failure_occ}, Graphonomy outputs a black patch (Background class) when an object or the scene is occluding the person. As \modelname-C tries to minimise the negative log probability of a rendered vertex being a particular label, and the background label has a low probability, occlusions can cause the pose to be incorrect.
More complete labeling of things like backpacks or training with synthetic occlusion could improve this.
Moreover, it can also hinder detailed fitting of the body where the labels associated with \modelname-MC are occluded. Additional occlusion handling techniques could help our approach in such cases.

Furthermore, another failure case occurs when multiple people are present in a scene. As Graphonomy is not an instance segmentation network, the pseudo ground-truth data may still contain other people even after using the heuristics to clean them, as described in Section~\ref{subsec:proc}. This confuses training, resulting in misaligned bodies at inference time. %
Figure \ref{fig:failure_multiperson} shows common cases where all the upper body clothing of multiple people are merged into one segment and clothing masks of partially visible people in the background, which affect the quality of the obtained masks.
Our entire method could be improved by better instance-level clothing segmentation.

Higher quality of Graphonomy masks leads to increased performance gains in the case of \modelname. We demonstrate it by doing an ablation study using Human3.6M~\cite{h36m_pami} dataset where the Graphonomy predictions are more reliable because of the simpler background and single subject. The quantitative results of this experiment are reported in the main paper.

Overall, our performance is affected by the off-the-self model we use to supervise the clothing semantics of the person. However, improvements over the state-of-the-art show that even weak supervision of clothing semantics is crucial for detailed 3D body fits.
The success of our approach suggests that more accurate human parsing and clothing segmentation are a good investment for the community.

\section{Additional Qualitative Results}
We show additional qualitative results comparing our method with other state-of-the-art methods~\cite{spin, eft} for 3DPW~\cite{Li20143DHP} and COCO~\cite{coco} which are challenging in-the-wild benchmarks for 3D human pose and shape estimation. The results are depicted in Figures~\ref{fig:additional_qual_3dpw} and \ref{fig:additional_qual_coco}. Next to each example, we show the corresponding side view. We observe that our approach produces more accurate pose and shape that are better aligned with the human in the image than current SOTA approaches.

\end{document}